\documentclass[11pt]{article}
\usepackage[utf8]{inputenc}

\usepackage{arabtex}
\usepackage{utf8}
\usepackage{graphicx}
\usepackage{tabularx}
\usepackage{soul}
\usepackage{booktabs}
\usepackage{makecell}
\usepackage{array}
\newcolumntype{C}{>{\centering\arraybackslash}p{4.7em}}
\usepackage{multirow}
\usepackage{colortbl}
\usepackage{comment}
\usepackage{wrapfig}

\usepackage{coling2020}
\usepackage{times}
\usepackage{url}
\usepackage{latexsym}

\DeclareFontFamily{U}{xnsh}{}%

\DeclareFontShape{U}{xnsh}{m}{n}{%
   <-6> sfixed * [6.0] xnsh14
      <6-10> s * [1.20] xnsh14
         <10><10.95><12><14.4><17.28><20.74><24.88> s * [1.20] xnsh14
         }{}

\DeclareFontShape{U}{xnsh}{bx}{n}{%
   <-6> sfixed * [6.0] xnsh14bf
   <6-10> s * [1.20] xnsh14bf
   <10><10.95><12><14.4><17.28><20.74><24.88> s * [1.20] xnsh14bf
}{}

\colingfinalcopy 

\title{AraWEAT:  Multidimensional Analysis of Biases \\ in Arabic Word Embeddings}

\author{\textbf{Anne Lauscher, Rafik Takieddin, Simone Paolo Ponzetto, and Goran Glava\v{s}}\\ 
Data and Web Science Research Group\\
University of Mannheim \\
\texttt{\{anne, simone, goran\}@informatik.uni-mannheim.de} \\
\texttt{rafik.takieddin@gmail.com} \\

}

\date{}

\begin{document}
\setcode{utf8}
\setarab
\maketitle
\begin{abstract}
Recent work has shown that distributional word vector spaces often encode human biases like sexism or racism. In this work, we conduct an extensive analysis of biases in Arabic word embeddings by applying a range of recently introduced bias tests on a variety of embedding spaces induced from corpora in Arabic. We measure the presence of biases across several dimensions, namely: embedding models (\textsc{Skip-Gram}, \textsc{CBOW}, and \textsc{FastText}) and vector sizes, types of text (encyclopedic text, and news vs.~user-generated content), dialects (Egyptian Arabic vs.\,Modern Standard Arabic), and time (diachronic analyses over corpora from different time periods). Our analysis yields several interesting findings, e.g., that implicit gender bias in embeddings trained on Arabic news corpora steadily increases over time (between 2007 and 2017). We make the Arabic bias specifications (AraWEAT) publicly available. 
\end{abstract}

\section{Introduction}
Recent research offered evidence that distributional word representations (i.e., word embeddings) induced from human-created text corpora exhibit a range of human biases, such as racism and sexism \cite{Bolukbasi:2016:MCP:3157382.3157584,Caliskan183}. With word embeddings ubiquitously used as input for (neural) natural language processing (NLP) models, this brings about the jeopardy of introducing stereotypical unfairness into NLP models, which can reinforce existing social injustices, and therefore be harmful in practical applications. 
For instance, consider the seminal gender bias example
\noindent\emph{``Man is to computer programmer as woman is to homemaker"}, which is algebraically encoded in the embedding space with the analogical relation
$
    \vec{\textit{man}}-\vec{\textit{computer programmer}}\approx \vec{\textit{woman}}-\vec{\textit{homemaker}}
$ \cite{Bolukbasi:2016:MCP:3157382.3157584}. The existence of such biases in word embeddings stems from the combination of (1) human biases manifesting themselves in terms of word co-occurrences (e.g., the word \emph{woman} appearing in a training corpus much more often in the context of \emph{homemaker} than together with \emph{computer programmer}) and (2) the distributional nature of the word embedding models \cite{mikolov2013distributed,pennington2014glove,Bojanowski:2017tacl}, which induce word vectors precisely by exploiting word co-occurrences, i.e., thus also encoding the human biases as a (negative) side-effect, which represents, expressed according to the taxnomy of harms proposed by \newcite{blodgett2020language}, a \emph{representational harm}, more specifically, \emph{stereotyping}.
%
In order to quantify the amount of bias in word embeddings, \newcite{Caliskan183} proposed the Word Embedding Association Test (\textsc{WEAT}), which is based on the associative difference in terms of semantic similarity between two sets of target terms, e.g., \emph{male} and \emph{female} terms, towards two sets of attribute terms, e.g., \emph{career} and \emph{family} terms. Most recently, the WEAT test, measuring the degree of explicit bias in the distributional space, has been coupled with other tests, aiming to measure other aspects of bias, such as the amount of implicit bias \cite{gonen2019lipstick} or the presence of the analogical bias \cite{lauscher2019general}. 

While there is evidence that distributional vectors often encode human biases, the amount of biases does not seem to be universal across different languages and corpora, as recently shown by \newcite{lauscher2019we} in the analysis of distributional biases across seven different languages. In this work, we focus on the multi-dimensional analysis of biases in Arabic word embeddings. The motivation for this work is twofold: (1) Arabic is one of the most widely spoken languages in the world:\footnote{According to \newcite{mikael2007varldens}, Arabic is the fifth most spoken language in the world with close to 300 million native speakers.} this means that the biases encoded in language technology for Arabic have the potential for affecting more people than for most other languages; (2) language resources for Arabic -- large corpora \cite{Goldhahn12buildinglarge}, pretrained word embeddings \cite{mohammad,Bojanowski:2017tacl}, and datasets for measuring semantic quality of Arabic embeddings \cite{elrazzaz-etal-2017-methodical,cer-etal-2017-semeval} -- are publicly available, allowing for the analyses of biases that these resources potentially hide.       


As a first step in the analysis of language technology biases for Arabic, we present \textsc{AraWEAT}, an Arabic extension to the multilingual \textsc{XWEAT} framework \cite{lauscher2019we}. Because the \textsc{WEAT} test \cite{Caliskan183}, despite being derived from an established test from psychology \cite{Nosek02harvestingimplicit}, has recently been shown to systematically overestimate the bias present in an embedding space \cite{Ethayarajh_2019}, in this work, we couple it with several other bias tests, designed to capture and quantify other aspects of human biases: Embedding Coherence Test \cite{dev2019attenuating}, Bias Analogy Test \cite{lauscher2019general} and Implicit Bias Tests \cite{gonen2019lipstick}.

Our work, which is to the best of our knowledge the first study on quantifying biases in Arabic distributional word vector spaces, yields some interesting findings: biases seem more prominent in vectors trained on texts written in Egyptian Arabic than those written in Modern Standard Arabic (MSA). Also, the implicit gender bias in Arabic news corpora seems to be steadily on the rise over the ten year period between 2007 and 2017. Finally, we find evidence that the explicit bias effects, as measured by the WEAT test, in embeddings trained on the entire Arabic news corpus roughly correspond to averaging the biases measured across embeddings trained on temporally disjunct subsets of the corpus.


\section{AraWEAT}
We present \textsc{AraWEAT}, our framework allowing for multi-dimensional analysis of bias in Arabic distributional word vector spaces.

\subsection{Data for Measuring Bias}

At the core of our extension are the Arabic bias test specifications, which are based on the original English \textsc{WEAT} test data. \textsc{WEAT} is an adaptation of the Implicit Association Test \cite{Nosek02harvestingimplicit}, which quantifies biases as association differences measured in terms of response times of human subjects when exposed to different sets of \emph{stimuli}. \textsc{WEAT}, in turn, measures the association differences in terms of the difference in semantic similarity between two sets of \emph{target} terms towards two sets of \emph{attribute} terms. 
\begin{wraptable}{r}{0.5\textwidth}
\setlength{\tabcolsep}{5pt}
\centering
\small{
{
\begin{tabular}{l | r}
\toprule
T1 & \multicolumn{1}{l}{\emph{math algebra geometry calculus equations computation}} \\ & \multicolumn{1}{l}{\emph{numbers addition}} \\
T2 & \multicolumn{1}{l}{\emph{poetry art dance literature novel symphony drama}}\\ & \multicolumn{1}{l}{\emph{sculpture}} \\
A1 & \multicolumn{1}{l}{\emph{male man boy brother he him his son}} \\
A2 & \multicolumn{1}{l}{\emph{female woman girl sister she her hers daughter}} \\
\midrule
T1 & \< معادلات	الرياضيات	الجبر	الهندسة	تحليل	إضافة	أعداد	حساب	>\ \\ 
T2 & \< الشعر	رقص	فن	الأدب	رواية	سمفونية	نحت	دراما >\ \\ 
A1 & \< له	ابن	صبي	الذكر	شقيق	رجل	هو	 >\ \\ 
A2 & \< ابنة	أخت	نساء	أنثى	فتاة	هي	لها	 >\ \\
\bottomrule
\end{tabular}
} }
\caption{Original English version and MSA translation of the WEAT Test 7 specification.}
\label{tbl:translation}
\end{wraptable}
Our creation of WEAT tests for Arabic starts with automatically translating, using Google Translate, the term sets (i.e., the terms from each of two target and two attribute lists) from the English WEAT tests. We then hired a native speaker of modern standard Arabic (MSA), who manually verified and, when needed, corrected the translations. Since Arabic is a language with grammatical genders, we made sure to account for both genders when translating the terms so that we do not artificially introduce a bias in our test specifications (e.g., we translated the genderless English word engineer as both \<مهندس> (\emph{engineer} m.) and \<مهندسة> (\emph{engineer} f.).  
While initially considered, we did not translate WEAT test specifications to the different Arabic dialects, as the differences between the MSA translations and dialectal translations for the terms from the WEAT test were observed only in a negligible fraction of cases; and even in those cases the MSA translation is also in usage in other Arabic dialects.\footnote{Although possibly less frequently than the dialectal translation.} 
We further omitted \textsc{WEAT} tests $3$--$6$ and $10$ because they are based on proper names. While it has been shown that names are a good proxy for identifying and removing bias towards specific groups of people \cite{hall-maudslay-etal-2019-name}, it is difficult to ``translate'' them.\footnote{Furthermore, \textsc{WEAT} tests $3$--$5$ are tailored to test racial biases towards African-Americans, which is arguably much less prominent in the Arabic cultural area.} As an example of the resulting AraWEAT test, Table~\ref{tbl:translation} list the Arabic translation of \textsc{WEAT} test $T7$. An overview on the remaining tests with their respective target and attribute term sets is provided in Table~\ref{tbl:weat}.
%
%
\setlength{\tabcolsep}{1.8pt}
\begin{table*}[t]
\centering
{\footnotesize
\begin{tabular}{c | lllll}
\toprule
\textbf{Test} & \textbf{Bias Type} & \textbf{Target Set \#1} & \textbf{Target Set \#2} & \textbf{Attribute Set \#1} & \textbf{Attribute Set \#2} \\ \midrule
1 & Universal & Flowers (e.g.,~\emph{aster})
& Insects (e.g.,~\emph{ant}, \emph{flea}) & Pleasant (e.g.,~\emph{health}) & Unpleasant~(e.g., \emph{abuse})
\\
2 & Militant &  Instruments (e.g.,~\emph{cello}) 
& Weapons (e.g.,~\emph{gun}) &
Pleasant & Unpleasant\\
7 & Gender & Math (e.g.,~\emph{algebra}, \emph{geometry}) & Arts (e.g.,~\emph{poetry}) & Male (e.g.,~\emph{brother}, \emph{son}) & Female (e.g.,~\emph{woman}, \emph{sister}) \\
8 & Gender & Science (e.g.,~\emph{experiment}) & Arts & Male & Female\\
9 & Disease & Physical (e.g.,~\emph{virus}) & Mental (e.g.,~\emph{sad}) & Long-term (e.g.,~\emph{always}) & Short-term (e.g.,~\emph{occasional})\\
\bottomrule
\end{tabular}
}
\caption{WEAT bias tests.}
\label{tbl:weat}
\end{table*}

\setlength{\tabcolsep}{1.5pt}
\begin{table*}[th!]
\centering
\small{
\begin{tabular}{l l | cccc | cccc | cccc  | cccc}
\toprule
& &\multicolumn{4}{c}{\textbf{T1}} & \multicolumn{4}{c}{\textbf{T2}} & \multicolumn{4}{c}{\textbf{T7}} & \multicolumn{4}{c}{\textbf{T8}} \\
\textbf{Model} & \textbf{Lang} & W & ECT & BAT & KM & W & ECT & BAT & KM & W & ECT & BAT & KM & W & ECT & BAT & KM \\
\midrule
\textsc{FT Arabic} & MSA & 0.85 & 0.69 & 0.47 & 0.71 & 0.51* & 0.62 & 0.43 & 0.63 & -0.15* & 0.17* & 0.5 & 0.56 & 0.05* & 0.02* & 0.44 & 0.56 \\
\textsc{FT Egypt} & Egyptian & 1.17 & 0.45 & 0.49 & 0.95 & 0.97 & 0.56 & 0.51 & 0.65 & 0.65* & 0.54 & 0.51 & 0.6 & 0.09* & 0.63 & 0.47 & 0.6\\
\midrule
\textsc{AV SG Wiki} & MSA & 0.27* & 0.82 & 0.49 & 0.62 & 0.98 & 0.61 & 0.43 & 0.92 & 0.22* & -0.6 & 0.57 & 0.88 & 0.13* & -0.53* & 0.64 & 0.72\\
\textsc{AV SG Twitter} & Mixed & 1.21 & 0.27* & 0.50 & 0.63 & 0.87 & 0.33 & 0.41 & 0.75 & 0.38* & 0.05* & 0.46 & 0.70 & -0.98* & 0.50* & 0.42 &  0.60 \\
\midrule
\textsc{AV CB Wiki} & MSA & 0.43* & 0.91 & 0.45 & 0.67 & 1.21 & 0.53 & 0.45 & 0.52 & 0.57* & -0.35* & 0.57 & 0.75 & -0.38* & 0.26* & 0.53 & 0.58\\
\textsc{AV CB Twitter} & Mixed & 1.00 & 0.53 & 0.39 & 0.78 & 0.92 & 0.54 & 0.43 & 0.71 & 0.41* & 0.48* & 0.31 & 0.72 & -0.49* & 0.83 & 0.40 & 0.76 \\
\bottomrule
\end{tabular}
}
\caption{Bias scores for 300-dimensional pretrained FastText (\textsc{FT}) and AraVec (\textsc{AV}) n-gram distributional word vector spaces. We omitted test $9$ as less than $20\%$ of the test vocabulary was found in the \textsc{FT} and \textsc{AV} embedding spaces. We report explicit bias scores in terms of WEAT (\textsc{W}), \textsc{ECT}, and \textsc{BAT}, and implicit bias in terms of KMeans++ accuracy (\textsc{KM}). For \textsc{AV}, we report results for the Skip-gram (\textsc{SG}) and CBOW (\textsc{CB}) models. Asterisks indicate WEAT bias effects or pearson correlation scores that are insignificant at $\alpha < 0.05$.}
\label{tbl:pretr}
\end{table*}
%
%
%
%
%
%
%
%
%
\setlength{\tabcolsep}{1.0pt}
\begin{table*}[t!]
\centering
\footnotesize{
\begin{tabular}{l | ccc ccc ccc ccc ccc ccc ccc ccc }
\toprule
 & \multicolumn{3}{c}{\textbf{$2007$}} & \multicolumn{3}{c}{\textbf{$2008$}} & \multicolumn{3}{c}{\textbf{$2009$}} & \multicolumn{3}{c}{\textbf{$2010$}} & \multicolumn{3}{c}{\textbf{$2011$}} & \multicolumn{3}{c}{\textbf{$2015$}} & \multicolumn{3}{c}{\textbf{$2016$}} & \multicolumn{3}{c}{\textbf{$2017$}} \\ 
\textbf{\# Sent} & W & KM & STS & W & KM & STS & W & KM & STS & W & KM & STS & W & KM & STS & W & KM & STS & W & KM & STS & W & KM & STS \\
\midrule
300K & -.35* & .75 & .38 & -.33* &  .76 & .38 & -.52* & .53 & .33 & -.45* & .50 & .34 & .06* & .64 & .35 & -.02* & .75  & .38 & .30* & .80 & .37 & .16* & .75 & .36 \\
1M & -- & -- & -- &  .08* & .53  & .44 & .92 & .55  & .40 & .32* & .58  & .38 & .50* & .65  & .37 & .97 & .71  & .42 & .72* & .71  & .43 & .95 & .72 & .42  
\\ 
\bottomrule

\end{tabular}
}
\caption{Gender bias over time: WEAT test 7 bias effects and KMeans++ accuracy scores for 300-dimensional distributional word vector spaces induced using CBOW on Leipzig news corpora of size consisting of 300k and 1M sentences between 2007 and 2017. Asterisks indicate bias effects that are insignificant at $\alpha < 0.05$.}
\label{tbl:timesize}
\end{table*}
\begin{table}
\setlength{\tabcolsep}{2pt}
\parbox{.46\linewidth}{
\centering
\small{
\begin{tabular}{l lcccc}
\toprule
\textbf{Model} & \textbf{Dim.} &\textbf{T1} & \textbf{T2} & \textbf{T7} & \textbf{T8} \\ 
\midrule
\textsc{AraVec SG unigram} & 100 &  0.04* & 0.91 & 0.66* & -0.50*\\
\textsc{AraVec SG unigram} & 300 & 0.18* & 0.66 & 0.13* & -0.19*\\
\midrule
\textsc{AraVec SG n-gram} & 100 & 0.49* & 1.19 & 0.54* & -0.49*\\
\textsc{AraVec SG n-gram} & 300 & 0.27* & 0.98 & 0.22* & 0.13* \\
\bottomrule
\end{tabular}
}
\caption{WEAT bias effect sizes for \textsc{AraVec Skip-gram} pretrained distributional word vector spaces trained on Wikipedia with embedding dimensionality $100$ vs. $300$ and unigram vs. n-gram preprocessing. Asterisks indicate the bias effects which are insignificant at $\alpha < 0.05$.}
\label{tbl:hyper}
}
%
%
%
%
%
%
%
\hfill
\parbox{.53\linewidth}{
\setlength{\tabcolsep}{3pt}
\centering
\small{
\begin{tabular}{l | cc cc cc cc | c}
\toprule
& \multicolumn{2}{c}{\textbf{T1}} & \multicolumn{2}{c}{\textbf{T2}} & \multicolumn{2}{c}{\textbf{T7}} & \multicolumn{2}{c}{\textbf{T8}} \\
& W & KM & W & KM & W & KM & W & KM & \textbf{STS}  \\
\midrule
\textsc{Avg} & 0.68 & 0.91 & 1.03 & 0.79 & 0.69 & 0.63 & -0.47 & 0.71 & 0.41\\
\textsc{Conc} & 0.65 & 0.75 & 1.28 & 0.68 & 0.56* & 0.72 &  -0.73 & 0.77 & 0.52 \\
\bottomrule

\end{tabular}
}
\caption{Explicit (WEAT, \textsc{W}) and implicit (K-Means++, \textsc{KM}) bias scores for 300-dim. embedding spaces induced using CBOW on the Arabic portions of the Leipzig News Corpora of 1M sentences between 2007 and 2017; comparison of averaged biases over temporally non-overlapping portions (\textsc{Avg}) with those in embeddings induced on the whole corpus (\textsc{Conc}). Asterisks: insignificant bias effects ($\alpha < 0.05$).}
\label{tbl:concat}
}
\end{table}
\subsection{Bias Evaluation Methodology}
Aiming towards a holistic picture of biases encoded in Arabic word vectors, we put together several bias tests that quantify both implicit and explicit biases: (1) WEAT \cite{Caliskan183}, (2) Embedding Coherence Test (\textsc{ECT}) \cite{dev2019attenuating}, (3) Bias Analogy Test (\textsc{BAT}) \cite{lauscher2019general}, and (4) Implicit Bias Test with K-Means++ (\textsc{KM}) \cite{gonen2019lipstick}. For all bias tests, we adopt the notion of \emph{implicit} and \emph{explicit} bias specifications as proposed by \newcite{lauscher2019general}: an explicit bias specification consists here of two sets of target terms and two sets of attribute terms $B_E(T_1, T_2, A_1, A_2)$. The idea is to measure the bias between the target sets, e.g., \emph{science} and \emph{art}, \emph{towards} the attribute sets, e.g., \emph{male} vs. \emph{female} terms, or vice versa. In contrast, an implicit bias specification consists of target terms only, i.e., $B_E(T_1, T_2)$. Accordingly, the intuition is to measure bias between the target term representations only, and not its explicit manifestation with regard to other concepts.
Furthermore, we report the semantic quality for all word embedding spaces we induced ourselves: to this end, we report the scores on predicting sentence-level semantic similarity for Arabic on the dataset from SemEval 2017 Task 1 \cite{cer-etal-2017-semeval}; we obtain sentence embeddings simply as averages of word embeddings \cite{cer-etal-2017-semeval,glavavs2018resource}.

\paragraph{Word Embedding Association Test (WEAT).}

Let $B_E(T_1, T_2, A_1, A_2)$ be an explicit bias specification consisting of two sets of \textit{target} terms $T_1$ and $T_2$, and two sets of \emph{attribute} terms, $A_1$ and $A_2$. \newcite{Caliskan183} define the \textsc{WEAT} test statistic $s(T_1, T_2, A_1, A_2)$ as the association difference that $T_1$ and $T_2$ exhibit w.r.t. $A_1$ and $A_2$ -- the association is measured as the average semantic similarity of $T_1$/$T_2$ terms with terms from $A_1$ and $A_2$: 

\small{
\begin{equation}
    s(T_1, T_2, A_1, A_2) = \sum_{t_1 \in T_1}{s(t_1, A_1, A_2)} - \sum_{t_2 \in T_2}{s(t_2, A_1, A_2)}\,, 
\end{equation}}

\normalsize
\noindent with associative difference for a term $t$ computed as: 

\small{
\begin{equation}
    s(t, A_1, A_2) = \frac{1}{|A_1|}\sum_{a_1 \in A_1}{\textnormal{cos}(\mathbf{t}, \mathbf{a_1})} - \frac{1}{|A_2|}\sum_{a_2 \in A_2}{\textnormal{cos}(\mathbf{t}, \mathbf{a_2})}\,, 
\end{equation}}

\normalsize
\noindent with $\mathbf{t}$ as the distributional vector of term $t$ and $\textnormal{cos}$ as the cosine of the angle between two vectors. The significance of the test statistic is measured by the non-parametric permutation test in which the $s(T_1, T_2, A_1, A_2)$ is compared to $s(X_1, X_2, A_1, A_2)$, where ($X_1$, $X_2$) denotes a random, equally-sized split of the terms in $T_1 \cup T_2$. A larger WEAT effect size indicates a larger bias.  





\paragraph{Embedding Coherence Test (ECT).} Given an \textsc{AraWEAT} explicit bias specification $B_E(T_1, T_2, A_1, A_2)$, ECT operates on the bias specification which ``collapses" the two AraWEAT attribute sets into a single set: $B_E(T_1, T_2, A = A_1 \cup A_2)$. Next, as proposed by \newcite{dev2019attenuating}, we compute the vectors $\mathbf{t_1}$ and $\mathbf{t_2}$ as averages of the word vectors of terms in $T_1$ and $T_2$, respectively. Then, we obtain two vectors of similarities by computing the \emph{cosine similarity} between the vector of each term in $A$ and the mean vectors $\mathbf{t_1}$ and $\mathbf{t_2}$. The ECT score is finally the Spearman correlation between the two obtained similarity vectors. The intuition is to assess, whether the similarities of the average vectors  $\mathbf{t_1}$ and $\mathbf{t_2}$, which represent the two target term sets, with the attribute terms are correlating. The larger the ECT correlation, the lower the bias. 

\paragraph{Bias Analogy Test (BAT).} Inspired by \newcite{Bolukbasi:2016:MCP:3157382.3157584}'s famous anology, the idea behind BAT is to quantify the fraction of biased analogies that result from querying the embedding space. Given an AraWEAT test $B_E(T_1, T_2, A_1, A_2)$, following \newcite{lauscher2019general}, we create all possible biased analogies $\mathbf{t}_1 - \mathbf{t}_2 \approx \mathbf{a}_1 - \mathbf{a}_2$ for $(t_1, t_2, a_1, a_2) \in T_1 \times T_2 \times A_1 \times A_2$. Next we create two query vectors -- $\mathbf{q}_1 = \mathbf{t}_1 - \mathbf{t}_2 + \mathbf{a}_2$ and $\mathbf{q}_2 = \mathbf{a}_1 - \mathbf{t}_1 + \mathbf{t}_2$ -- for each tuple $(t_1, t_2, a_1, a_2)$. We then rank the vectors in the vector space according to the Euclidean distance with $q_1$ and $q_2$, respectively, and report the percentage of cases where: $a_1$ is ranked higher than a term $a'_2 \in A_2\setminus\{a_2\}$ for $\mathbf{q}_1$ and $a_2$ is ranked higher than a term $a'_1 \in A_1\setminus\{a_1\}$ for $\mathbf{q}_2$. The higher the BAT score, the higher the bias. 

\paragraph{Implicit Bias Test: K-Means++ (KM).} Sometimes, bias is not expressed explicitely, i.e., as bias between two target term sets in explicit relation towards certain attribute sets, but manifests implicitly. In order to additionally reflect this type of bias in our study, we follow \newcite{gonen2019lipstick} and test the Arabic word vector spaces for the amount of implicit bias by clustering terms from $T_1$ and $T_2$ with KMeans++ \cite{Arthur2007kmeans}. The higher the clustering accuracy, the higher the bias. We report the averaged accuracy over $20$ independent runs.

\paragraph{Semantic Quality (SQ).}

For the embedding models we train ourselves, we additionally report the semantic quality of the space by predicting sentence-level semantic similarity on the SemEval 2017 Task 1 for Arabic (ar-ar) \cite{cer-etal-2017-semeval}. 
Let $\mathbf{s_a}={e_{a1}, ..., e_{an}}$ be the set of embeddings of words in sentence $a$ and let $\mathbf{s_b}={e_{b1}, ..., e_{am}}$ be the sequence of embedding representations for individual words in sentence $b$. We obtain aggregated sentence representations, by averaging the embeddings of words in the sentence: $\mathbf{s}=\frac{1}{l}\sum_{i=1}^{l}e_i$ and finally predict the similarity score as $cos(\mathbf{s}_a, \mathbf{s}_b)$.\footnote{This method was used as the simple aggregation baseline in the corresponding SemEval shared task.} We report Pearson correlation between our predicitions and the gold similarity annotations.

\paragraph{Dimensions of Bias Analysis.}
We run our tests along $5$ different dimensions: (1) embedding methods: we compare embeddings induced using \textsc{Skip-Gram}, \textsc{CBOW} and \textsc{FastText} embedding models; (2) source text types: we analyze vector spaces induced from corpora originating from different sources (Wikipedia, news, Twitter);\footnote{While Arabic Wikipedia is dominantly written in MSA, \textsc{Twitter} is likely to exhibit non-negligible amounts of dialectical and colloquial Arabic.} (3) vector sizes and preprocessing: we hypothesize that biases might be more prominent in higher-dimensional vectors. To this end, we compare $100$- vs. $300$-dimensional embeddings. Furthermore, we analyze the effect of unigram vs. n-gram preprocessing of Arabic text, as offered by pretrained vectors AraVec \cite{mohammad}; (4) corpus size: \newcite{lauscher2019we} hypothesize that biases might be more expressed in bigger corpora. To further investigate this, we run several experiments controlling for corpus size; (5) temporal intervals: lastly, we conduct a diachronic bias analysis by training embeddings on corpora from different time periods.

\paragraph{Distributional Word Vector Spaces.}

We conduct our analysis on (a) pretrained distributional word vector spaces from AraVec\footnote{\url{https://github.com/bakrianoo/aravec}} \cite{mohammad} and FastText\footnote{\url{https://dl.fbaipublicfiles.com/fasttext/vectors-crawl/cc.ar.300.vec.gz}} \cite{Bojanowski:2017tacl} and (b) embedding spaces we trained in order to be able to control for corpora size and preprocessing. In (b), we use Arabic corpora from the Leipzig Corpora Collection\footnote{\url{http://wortschatz.uni-leipzig.de/en/download/}} \cite{Goldhahn12buildinglarge}.
\section{Findings}
We present and discuss the findings of our analysis employing \textsc{AraWEAT}.

\paragraph{Embedding Methods, Text Sources, and Dialects.}

Bias scores for 300-dimensional pretrained Fasttext (\textsc{FT}) and AraVec (\textsc{AV}) embedding spaces are shown in Table~\ref{tbl:pretr}. For both (\textsc{FT}) and (\textsc{AV}), we evaluated all available spaces, pretrained on different corpora. For \textsc{FT}, we investigate two models, one trained on the portions of Wikipedia and CommonCrawl corpora written in Modern Standard Arabic (MS) and the other on portions written in \emph{Egyptian Arabic}.\footnote{The language identification was performed automatically using the \textsc{FT} Language Detector} We evaluate the four variants of \textsc{AraVec} vectors: (a) trained using either Skip-Gram (\textsc{SG}) or CBOW (\textsc{CB}) on (b) either Wikipedia (\textsc{Wiki}) or Twitter (\textsc{Twitter}) text.
Interestingly, most of these embedding spaces fail to exhibit significant explicit gender biases according to WEAT tests $T7$ and $T8$. However, the gender biases seem to be rather present implicitly (KM) in most spaces. Comparing \textsc{FT Arabic} versus \textsc{FT Egyptian}, both implicit and explicit bias seems to be slightly more pronounced in the Egyptian than in the MSA corpus. Results of comparison over text types support the unexpected finding for other languages \cite{lauscher2019we}: embeddings built from user-generated content on average do not encode more bias than their counterparts trained on Wikipedia.


\paragraph{Embedding dimensionality and preprocessing.} Next, we evaluate the effects of specific hyperparameter settings using the \textsc{AraVec} pretrained vector spaces. \textsc{AraWEAT} bias effect sizes for different embedding dimensionalities and model types  are listed in Table \ref{tbl:hyper}.
For the \textsc{AraWEAT} test specifications $T1$, $T7$, and $T8$, we did not observe prominent variance in the amount of explicit bias w.r.t. the vector dimensionality or pre-processing type. For the remaining test -- $T2$ -- the explicit bias (according to the WEAT test) is somewhat more pronounced in the lower-dimensional embeddings and in the n-gram versions of the AraVec embeddings.

\paragraph{Diachronic Analysis and Corpora Sizes.} Table \ref{tbl:timesize} displays WEAT effect sizes for test $T7$ (\emph{gender bias}) in MSA 300-dimensional distributional word vector spaces we trained on the (temporally) disjunctive Arabic portions of the Leipzig News Corpora of sizes 300K and 1M sentences, respectively. 

%
%

The smaller corpus, consisting of $300$K sentences, exhibits no significant bias effect sizes across all years. This finding is in line with previous observations \newcite{lauscher2019we} that biases might be more expressed in embedding spaces obtained on bigger corpora. This could be a reflection of the overall quality of distributional vectors, which is lower when vectors are trained on smaller corpora (as supported by the corresponding \textsc{STS} scores). In the spaces obtained on the larger corpora segments, consisting of $1$M-sentences, significant explicit (W) gender biases are present in years $2009$, $2015$, and $2017$, with very similar effect sizes (between $.92$ and $.97$). The implicit gender bias (KM), on the other hand, steadily rises over the entire period under investigation (2007-2017).

Finally, we investigate how the biases in the embedding space induced on the whole Arabic Leipzig News corpus (2007--2017, \textsc{CONC}) relates to the biases detected in embedding spaces induced from its different, temporally non-overlapping subportions. To this end, we average the biases measured on embeddings trained on its yearly subsets (\textsc{AVG}). The correlation results, over all four tests and two measures (W, KM), are shown in Table~\ref{tbl:concat}. Indeed, the biases of the whole corpus (\textsc{CONC}) seem to be highly correlated with the averages of biases of subcorpora (\textsc{AVG}): we measure a substantial Pearson correlation of $66\%$ between the two sets of scores (\textsc{AVG} and \textsc{CONC}). This would suggest that one can roughly predict the biases of (embeddings trained on) a large corpus by aggregating the biases of (embeddings trained on) its (non-overlapping) subsets.  


\section{Related Work}
\newcite{Bolukbasi:2016:MCP:3157382.3157584} were the first to study bias in distributional word vector spaces. Using an analogy test, they demonstrate gender stereotypes manifesting in word embeddings and propose the notion of the bias direction, upon which they base a debiasing method called hard-debiasing.
\newcite{Caliskan183} adapt the Implicit Association Test (IAT) \cite{Nosek02harvestingimplicit} from psychology for studying biases in distributional word vector spaces. The test, dubbed Word Embedding Association Test (WEAT), measures associations between words in an embedding space in terms of cosine similarity between the vectors. They propose $10$ stimuli sets, which we adapt in our work. Later, \newcite{mccurdy2018} extend the analysis to three more languages, Dutch, German, and Spanish, but only focus on gender bias. XWEAT, the cross-lingual and multilingual WEAT framework \cite{lauscher2019we}, covers German, Spanish, Italian, Russian, Croatian, and Turkish. 
XWEAT analyses also focused on other relevant dimensions such as embedding method and similarity measures. \newcite{zhou2019examining} focus on measuring bias in languages with grammatical gender.
Several research efforts produced new bias tests: \newcite{dev2019attenuating} propose the Embedding Coherence Test (ECT) with the intuition of capturing whether two sets of target terms are coherently distant from a set of attribute terms. They also propose several debiasing methods.
\newcite{gonen2019lipstick} show that many debiasing methods only mask but do not fully remove biases present in the embedding spaces. They propose to additionally test for implicit biases, by trying to classify or cluster the sets of target terms. \newcite{lauscher2019general} unify the different notions of biases into explicit and implicit bias specifications, based on which they propose methods for quantifying and removing biases. 
While their is some effort to account for gender-awareness in Arabic machine translation~\cite{habash-etal-2019-automatic}, we are, to the best of our knowledge, the first to measure bias in Arabic Language Technology.
\section{Conclusion}
Language technologies should aim to avoid reflecting negative human biases such as racism and sexism. Yet, the ubiquitous word embeddings, used as input for many NLP models, seem to encode many such biases. In this work, we extensively quantify and analyze the biases in different vector spaces built from text in Arabic, a major world language with close to 300M native speakers. 
To this effect, we translate existing bias specifications from English to Arabic and investigate biases in embedding spaces that differ over several dimensions of analysis: embedding models, corpora sizes, type of text, dialectal vs. standard Arabic, and time periods. Our analysis yields interesting results. First, we confirm some of the previous findings for other languages, e.g., that biases are generally not more pronounced in user-generated text and that embeddings trained on larger corpora lead to more prominent biases. Secondly, our results suggest more bias is present in dialectal (Egyptian) Arabic corpora than in Modern Standard Arabic corpora.
Next, our diachronic analysis suggests that the implicit gender bias of Arabic news text steadily increases over time. Finally, we show that the bias effects of the whole corpus can be predicted from bias effects of its subcorpora.
We hope that \textsc{AraWEAT}, our framework for multidimensional analysis of stereotypical bias in Arabic text representations, fuels more research on bias in Arabic language technology. 


%
%
\blfootnote{
    %
    %
    %
    %
    %
    %
    \hspace{-0.65cm}  
     This work is licensed under a Creative Commons 
     Attribution 4.0 International License.
     License details:
     \url{http://creativecommons.org/licenses/by/4.0/}.
}

\section*{Acknowledgments}
Anne Lauscher and Goran Glavaš are supported by the Eliteprogramm of the Baden-Württemberg Stiftung (AGREE grant). We would like to thank the anonymous reviewers for their helpful comments.

\bibliographystyle{coling}
\bibliography{coling2020}

\end{document}